\documentclass[sigconf]{acmart}

\AtBeginDocument{%
  \providecommand\BibTeX{{%
    \normalfont B\kern-0.5em{\scshape i\kern-0.25em b}\kern-0.8em\TeX}}}
\usepackage{algorithm}
\usepackage{algorithmic}
\usepackage{booktabs}
\usepackage{makecell}
\usepackage{multirow}

\copyrightyear{2021}
\acmYear{2021}
\setcopyright{acmcopyright}\acmConference[MM '21]{Proceedings of the 29th ACM International Conference on Multimedia}{October 20--24, 2021}{Virtual Event, China}
\acmBooktitle{Proceedings of the 29th ACM International Conference on Multimedia (MM '21), October 20--24, 2021, Virtual Event, China}
\acmPrice{15.00}
\acmDOI{10.1145/3474085.3475377}
\acmISBN{978-1-4503-8651-7/21/10}

\newcommand\blfootnote[1]{% 
\begingroup 
\renewcommand\thefootnote{}\footnote{#1}% 
\addtocounter{footnote}{-1}% 
\endgroup 
}

\begin{document}
\fancyhead{}
\title{Coarse to Fine: Domain Adaptive Crowd Counting via Adversarial Scoring Network}


\author{Zhikang Zou$^\dagger$}
\affiliation{%
  \institution{The Hubei Engineering Research Center on Big Data Security, School of Cyber Science and Engineering, Huazhong University of Science and Technology, China
  \\ \& Department of Computer Vision Technology (VIS), Baidu Inc., China}}
\email{zouzhikang@baidu.com}

\author{Xiaoye Qu$^\dagger$}
\affiliation{%
  \institution{The Hubei Engineering Research Center on Big Data Security, School of Cyber Science and Engineering, Huazhong University of Science and Technology, China}}
\email{xiaoye@hust.edu.cn}

\author{Pan Zhou$^*$}
\affiliation{%
  \institution{The Hubei Engineering Research Center on Big Data Security, School of Cyber Science and Engineering, Huazhong University of Science and Technology, China}}
\email{panzhou@hust.edu.cn}

\author{Shuangjie Xu}
\affiliation{%
  \institution{Department of Computer Science and Engineering, Hong Kong University of Science and Technology, China}}
\email{shuangjiexu@gmail.com}

\author{Xiaoqing Ye}
\affiliation{%
  \institution{Department of Computer Vision Technology (VIS), Baidu Inc., China}}
\email{yexiaoqing@baidu.com}

\author{Wenhao Wu}
\affiliation{%
  \institution{Department of Computer Vision Technology (VIS), Baidu Inc., China}}
\email{wuwenhao01@baidu.com}

\author{Jin Ye}
\affiliation{%
  \institution{Department of Computer Vision Technology (VIS), Baidu Inc., China}}
\email{yejin16@mails.ucas.ac.cn}


\begin{abstract}
 Recent deep networks have convincingly demonstrated high capability in crowd counting, which is a critical task attracting widespread attention due to its various industrial applications. Despite such progress, trained data-dependent models usually can not generalize well to unseen scenarios because of the inherent domain shift. To facilitate this issue, this paper proposes a novel adversarial scoring network (ASNet) to gradually bridge the gap across domains from coarse to fine granularity. In specific, at the coarse-grained stage, we design a dual-discriminator strategy to adapt source domain to be close to the targets from the perspectives of both global and local feature space via adversarial learning. The distributions between two domains can thus be aligned roughly. At the fine-grained stage, we explore the transferability of source characteristics by scoring how similar the source samples are to target ones from multiple levels based on generative probability derived from coarse stage. Guided by these hierarchical scores, the transferable source features are properly selected to enhance the knowledge transfer during the adaptation process. With the coarse-to-fine design, the generalization bottleneck induced from the domain discrepancy can be effectively alleviated. Three sets of migration experiments show that the proposed methods achieve state-of-the-art counting performance compared with major unsupervised methods.
\end{abstract}





\maketitle

\blfootnote{$^\dagger$Equal Contribution.}
\blfootnote{$^*$Corresponding author: Pan Zhou.}

\section{Introduction}
Crowd counting is a core task in computer vision, which aims to estimate the number of pedestrians in a still image or video frame. In the last few decades, researchers have devoted significant efforts to this area and achieved remarkable progress in promoting the performance on the existing mainstream benchmark datasets. However, training convolutional neural networks requires large-scale and high-quality labeled datasets, while annotating pixel-level pedestrian locations is prohibitively expensive. Moreover, models trained on a label-rich data domain (source domain) can not generalize well to another label-scare domain (target domain) due to the domain shift among data distribution, which severely limits the practical applications of the existing methods. 

\begin{figure}[t]   
\centering
\includegraphics[width=\linewidth]{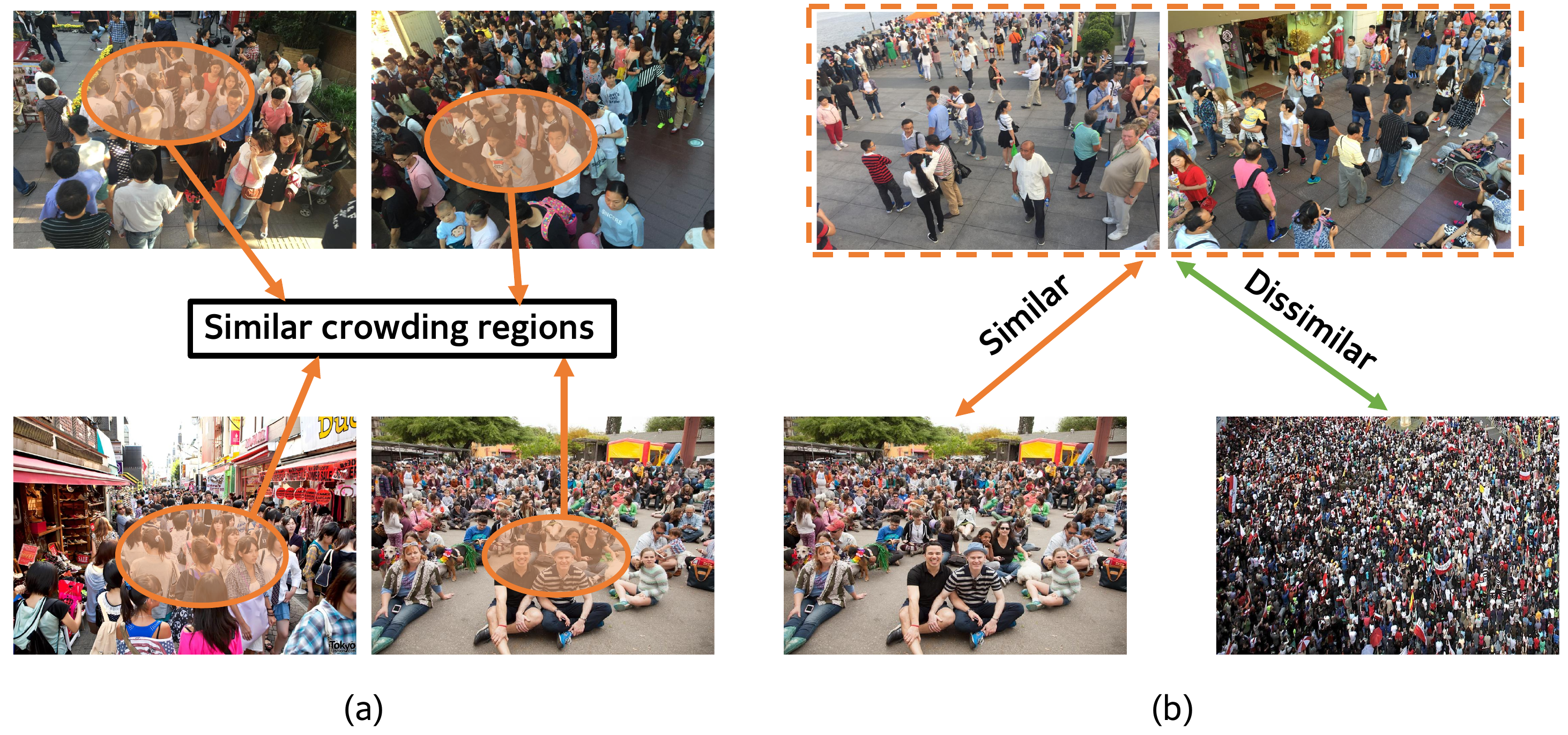}
\caption{Illustration of the domain similarity existing between the target (top) and the source domains (bottom). Left: some crowding regions are similar across domains in the pixel level. Right: partial source samples may share similar image distribution with target samples.}
\label{introduction}
\end{figure}

To alleviate the issues caused by the domain gap, a technique named unsupervised domain adaptation (UDA) has been preliminarily explored for crowd counting \cite{li2019coda,gao2019domain,gao2019feature,liu2021leveraging}. The key point of UDA is to make use of a domain discriminator to classify patches into the source or the target domains while the deep learner tries to confuse the discriminator in an adversarial manner. In this way, domain-invariant feature representations can be learned to bridge the source and the target data distributions. Typically, CODA \cite{li2019coda} combines the adversarial learning with self-supervised ranking loss to close the gap between the source and the target domains, where the estimation errors on the target samples can be reduced significantly. Differently, DACC \cite{gao2019domain} is proposed to translate the synthetic data to realistic images in a style transfer framework. Thus, models trained on translated data can be applied to the real world. Despite the dramatic performance improvement in reducing the domain discrepancy, these methods suffer from solely harnessing the whole natural characteristics of the source domain without further considering the similarity between the source domain and target domain. For crowd counting, some regions such as cluster background are not fit to transfer while equally treating all regions may lead to negative transfer. Besides, some source images are far from similar with the target domains, resulting in weak transferability. As shown in Figure \ref{introduction}, the similarity across domains appears in both pixel level and image level as images from two domains share similar crowding regions and implicit distribution. Therefore, it is essential to adapt patterns from different granularities according to their actual contributions for the joint learning.

In this paper, we propose a novel coarse-to-fine framework named Adversarial Scoring Network (ASNet) for domain adaptive crowd counting. At the coarse stage, we design a dual-discriminator strategy to conduct distribution alignment between the source and the target domains. Different from previous methods that merely focus on the whole image, this strategy also draws patches closer in the feature space. Specifically, the proposed strategy is composed of two parts: a global discriminator takes a whole image as input and a local discriminator accepts patches. Through adversarial learning, the domain discrepancy can be reduced from different perspectives. At the fine-grained stage, we explore the variability of transferability in the source domain depending on the fact that images or pixels with different feature distances from the target domain will contribute to the model generalization at varying degrees. Based on the output probabilities from two discriminators, we can produce significance-aware scores of multiple granularities: 1) image level; 2) patch level; 3) pixel level and 4) patch-pixel level. The scores of the image level and the patch level are generated to indicate the overall similarity of an image or specific image patch since images or patches more similar across domains are fit for the distribution alignment. Meanwhile, for the pixel-wise counting task, it is likely to find domain similarity in the pixel region. Therefore, the pixel level and the patch-pixel level scores precisely evaluate the similarity for images or input patches between two domains pixel by pixel. These scores are utilized to guide the density map learning in the source domain, which enhances the attention on transferable features among the domains during the adaptation process and thus promotes the adaptation accuracy without additional computational cost during inference. In experiments, we conduct three sets of adaptations and the results demonstrate that our model achieves state-of-the-art performance over the existing mainstream methods. Although unsupervised, it is worth noting that our model can obtain comparable results with fully-supervised models trained on the target dataset. 

To summarize, the main contributions are as follows:
\begin{itemize}
\setlength{\itemsep}{0pt}
\setlength{\parsep}{0pt}
\setlength{\parskip}{0pt}
    \item To the best of our knowledge, it’s the first attempt to implement fine-grained transfer in crowd counting by exploiting valid knowledge existing in the source domain from various granularities.
    \item We propose a novel adversarial scoring network (ASNet) to generate significance-aware scores at different levels and utilize these scores as the supervisory signal to guide density optimization in a self-learning manner, which can boost the performance of adaptation in crowd counting.
    \item Extensive migration experiments indicate that ASNet can achieve state-of-the-art performance, which demonstrates the effectiveness of the proposed approach.
\end{itemize}

\begin{figure*}[t]
\centering
\includegraphics[width=\linewidth]{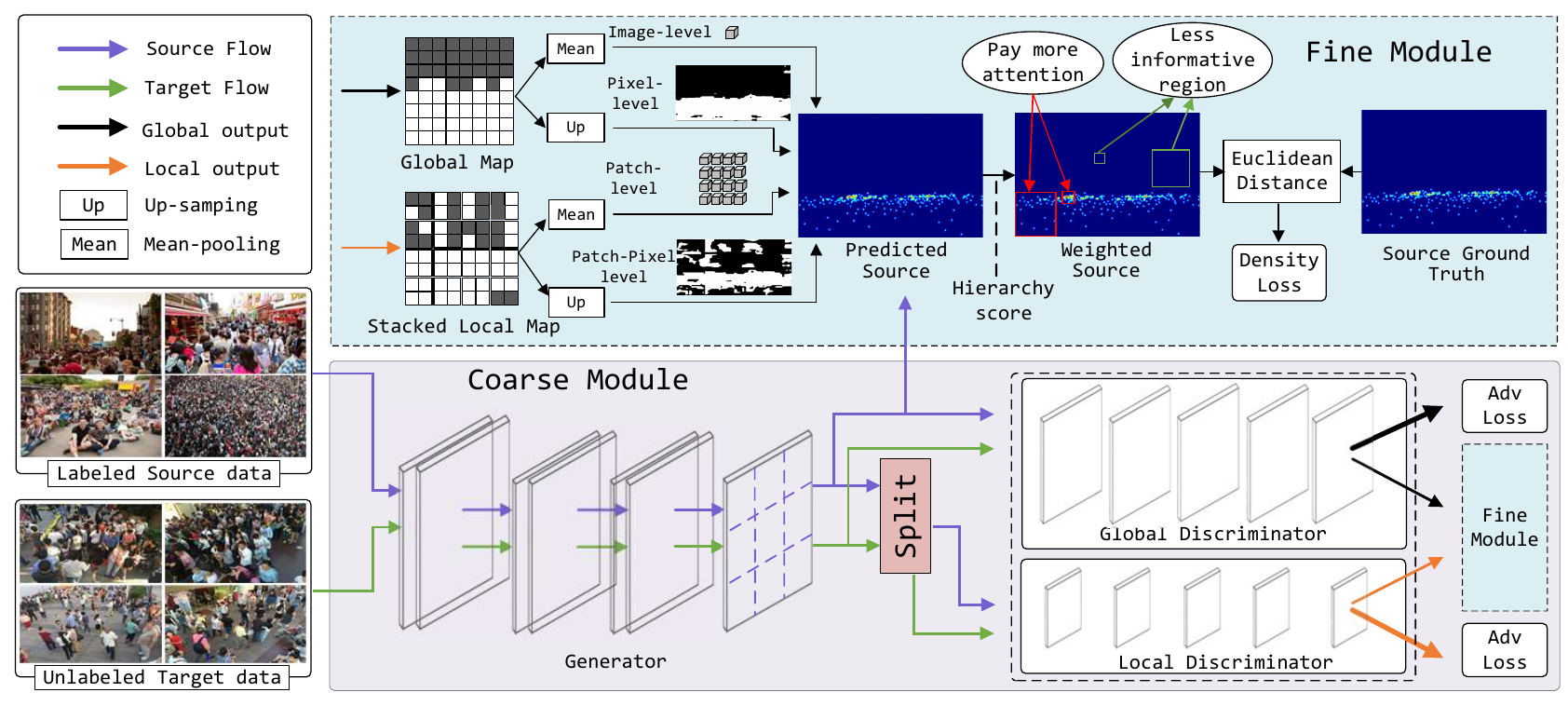}
\caption{Overview of our adversarial scoring network (ASNet). The generator encodes input images into density maps. Then the dual-discriminator classifies the density maps into source or target domain. By adversarial training between the generator and dual-discriminator, domain distributions are drawn close. Meanwhile, the dual-discriminator further produces four types of scores as a signal to guide the density optimization of source data, which achieves fine-grained transfer during adaptation.}
\label{whole_structure}
\end{figure*}

\section{Related Work}
\noindent\textbf{Crowd Counting.} Early works of crowd counting focus on detection style framework \cite{leibe2005pedestrian,li2008estimating,5975165}, where the body or part-based detectors are deployed to localize and count the pedestrians. These detection-based algorithms are limited by severe occlusions and complex background clusters in high-density crowd scenes. Hence, regression-based alternatives \cite{chan2009bayesian,ryan2009crowd} are proposed to directly estimate the number of people by learning a mapping from image features to the count number. The success of methods in this category lies in their ability to evade explicit detection. Nevertheless, these regression-based methods lose localization capability, which will lead to a performance drop as spatial awareness is totally ignored. The recent focus in counting area has been towards exploiting the advances in Convolutional Neural Networks (CNNs) \cite{cao2018scale,idrees2018composition,laradji2018blobs,abousamra2020localization,he2021error,liu2020cross,8497050} due to the remarkable representation learning ability. Typically, the majority of existing CNN-based frameworks are centered on coping with the large variation in pedestrian scales via the deployment of different architectures including multi-column networks \cite{zhang2016single,sam2017switching}, context-aware designs \cite{sindagi2017generating,liu2019context}, multi-task learning \cite{huang2017body,kang2018beyond} and others \cite{li2018csrnet,zhang2019attentional,wan2019adaptive}. In particular, Zhang \textit{et al.} \cite{zhang2016single} employ a multi-column convolutional network (MCNN) that captures diverse crowd scales with different receptive fields in each column. More recently, CSRNet \cite{li2018csrnet} connects VGG16 with dilated kernels to fuse multi-scale contextual information. In this way, the scale space captured by the model can be enlarged without a significant increase in computation. DM-Count \cite{wang2020distribution} measures the similarity between the normalized predicted density map and the normalized ground truth density map and designs OT loss and TV loss to boost the performance. SASNet \cite{song2021choose} automatically learns the internal correspondence between the scales and the feature levels. All these methods demonstrate the superiority of the convolutional neural networks.

\noindent\textbf{Domain adaptation.} To address the domain shift problem, technique named unsupervised domain adaptation (UDA) has been introduced for many computer vision tasks such as image classification \cite{geng2011daml,ganin2014unsupervised}, person re-identification \cite{bak2018domain,fu2019self}, semantic segmentation \cite{zou2018unsupervised} and other tasks \cite{hu2019multi, hu2020discriminative}. These methods aim to mitigate the domain gap and generalize the model onto the different test domain. To the best of our knowledge, there are very few domain adaptation frameworks \cite{li2019coda,wang2019learning,gao2019domain,gao2019feature,han2020focus} proposed for crowd counting area. They can be summarized into two widely used strategies: distribution alignment and image translation. The former \cite{li2019coda} 
adopts a discriminator to distinguish between the density maps generated by source image patches and targets. Thus, the data distributions across domains can be drawn closer. Differently, the latter \cite{wang2019learning} translates the synthetic labeled images to be similar to realistic scenes via a SE Cycle GAN. In this way, they train a CNN on translated images and obtain a reasonable result on real datasets. Despite the promising results, one common drawback of these methods is that they merely reduce the domain gap in a global view while ignoring the fine-grained transferability of source samples. 

% In this paper, we score the source samples from different aspects according to how similar these samples are close to the target domain. By generating scores of different levels, our model can adaptively focus on transferable pixels or images in the source domain, thus significantly boosting the adaptation accuracy.

\section{methodology}
In this section, we will introduce our proposed adversarial scoring network (ASNet). The goal is to improve the performance of crowd counting on the target domain by domain adaptation. Our core idea is to score how similar the source samples are to target ones from multiple levels and enhance the knowledge transfer guided by these hierarchical scores during the adaptation process. In specific, our ASNet consists of two parts: coarse module is designed to align the feature representation space across domains in global and local view, and fine module digs into the transferable samples of the source domain during coarse align process and generate guidance weights to further narrow the distance between domains. The overall pipeline is depicted in Figure \ref{whole_structure}. 

\subsection{Problem Formulation}
We have source domain data $X_s={\{(x_i^s, y_i^s)\}}_{i=1}^{N_s}$, where $x_i^s$ denotes the input source RGB image, $y_i^s$ represents its corresponding real-valued density map, and $N_s$ is the number of source domain labeled samples. Similarly, we have target domain unlabeled data $X_t={\{(x_i^t)\}}_{i=1}^{N_t}$. Here the source domain and target domain share different distribution, which appears in separate image background or crowd density. During training, we use both labeled source data and unlabeled target data as network input. 

\subsection{Coarse Adaptation}
The main difficulty of domain adaptation is the domain shift between the source domain and the target domain. Thus, it is important to reduce domain discrepancy during the training stage. Meanwhile, guaranteeing the quality of the predicted density map on the target domain is also fundamental. To achieve these goals, we consider performing density adaptation which minimizes the distance between density maps from two domains. 

To realize density adaptation, an intuitive idea based on generative adversarial network \cite{goodfellow2014generative} is adopted. The main principle is a two-player game, in which the first player is a domain discriminator $D$ whose role is to correctly classify which domain the features come from, while the second is a feature generator $G$ who aims to deceive the domain discriminator. In our task, the discriminator takes responsibility for distinguishing between the density maps generated by the source image and the target image. To capture a wider perspective in complicated crowd scenes, we utilize a dual-discriminator strategy as shown in Figure \ref{whole_structure}, namely global discriminator and local discriminator. The global discriminator $D_1$ takes the whole density map as input and the local discriminator $D_2$ accepts patches of density map. Then they output discrimination maps $O_1$ and $O_2$ in which each point value is normalized into [0,1] by sigmoid function, corresponding to confidence score belong to the source domain or the target domain. For both discriminators, binary cross-entropy loss is used to measure classification error. In specific, the loss can be formulated as:
\begin{small}
\begin{equation}
L_{d1} = -\frac{1}{N_b}(\sum_{i=1}^{N_b}\sum_{k=1}^{N_{p1}}log({(O_{1}^s})_{ik})-\sum_{i=1}^{N_b}\sum_{k=1}^{N_{p1}}log(1-{(O_{1}^t})_{ik}))
\end{equation}
\end{small}
\vspace{-5px}
\begin{small}
\begin{equation}
L_{d2} = -\frac{1}{N_b}(\sum_{i=1}^{N_b}\sum_{j=1}^{S^2}\sum_{k=1}^{N_{p2}}log((O_{2}^s)_{ijk})-\sum_{i=1}^{N_b}\sum_{j=1}^{S^2}\sum_{k=1}^{N_{p2}}log(1-(O_{2}^t)_{ijk}))
\end{equation}
\end{small}
where $O_1=D_1(G(x))$, $O_2=D_2(G(x))$, $N_{p1}={H_1}\times{W_1}$ is total pixel number of $O_1$, $N_{p2}={H_2}\times{W_2}$ is pixel number of $O_2$, $N_b$ is the number of training batch size, $S^2$ is the number of patches which are equally split from the density map.

To make density maps generated from source and target domain are more similar, we adopt an adversarial loss to guide the optimization of generator which further produces density maps to fool the discriminator. At the same time, considering that images from different domains may share local similarity, the generator also demands to generate similar patches of density map. The adversarial losses corresponding to two discriminators are:
\begin{align}
L_{adv1} &= -\frac{1}{N_b}\sum_{i=1}^{N_b}\sum_{k=1}^{N_{p1}}log((O_{1}^t)_{ik}) \\
L_{adv2} &= -\frac{1}{N_b}\sum_{i=1}^{N_b}\sum_{j=1}^{S^2}\sum_{k=1}^{N_{p2}}log((O_{2}^t)_{ijk})
\end{align}
It is worth noting that we only compute adversarial loss on target images for the generator.

\subsection{Fine Adaptation}
With the global density adaptation, the domain gap between the source and target domains is reduced. However, the above adaptation mainly aligns the images from a global perspective. It neglects the fact that not all regions of an image are suitable for transfer, such as the background which shows a significant difference between the two domains. Meanwhile, partial images in the source domain are more similar to target domain images than other parts. Thus, it is essential to pay attention to each pixel in an image and each image in the source domain instead of treating all images equally. To this end, we propose a fine-grained adaptation to achieve the pixel and image knowledge transfer. In specific, we define four scoring levels for source images from coarse to fine: 1) image level $W_1$; 2) patch level $W_2$; 3) pixel level $W_3$ and 4) patch-pixel level $W_4$. Here the image-level and patch-level scores determine the transferability of a complete source image or patches of the image corresponding to target image, and it is reasonable to give more focus on those source images with similar distribution to the target images. The pixel level and patch-pixel level scores measure the similarity pixel by pixel between source and target images or patches. Hence, it is also useful to favor the regions from a source image that are highly similar to the target ones. 

As mentioned above, our aim is to score the whole image and regions in each image from the source domain. In order to obtain the scores, we utilize the outputs of two discriminators which are the probability of the input belonging to the source domain. For the global discriminator, the output approaching 1 indicates the input image belongs to the source domain. Similarly, the input patches belong to the source domain if the output of the local discriminator is close to 1. For the output of the global discriminator $O_1^s$, we perform average pooling to obtain domain probability. Then we set threshold to obtain the image level score $W_1$. This process can be formulated as:
\begin{equation}
M_i = \text{Average}((O_{1}^s)_i), \quad W_1 = I(M_i<0.5) 
\end{equation}
where $(O_{1}^s)_i$ is the global discriminator output corresponding to input image $x_i^s$, $I(\cdot)$ is an indication function. $W_1$ is a binary scalar which denotes the transferability of the whole image. 
Meanwhile, we can get the pixel level score $W_3$ from the discriminator output. However, the size of the output discrimination map is not compatible with the input image. Thus, we conduct the nearest up-sampling and soft threshold to obtain the pixel level score:
\begin{equation}
P_i = \text{Up-sample}((O_{1}^s)_i), \quad W_3 = I(P_i<\text{mean}(P_i)) 
\end{equation}
where each value of $W_3$ denotes the similarity of the corresponding point of source image. The soft threshold uses the mean value as the threshold which can adapt to various score ranges and guarantee to select some relatively similar regions compared to a hard threshold.  
In the same way, with the output of local discriminator, we can get the patch level score $W_2$ and patch-pixel level score $W_4$ from the local discriminator output. 

After getting four fine-grained scores, we weight the density loss for the source domain. Formally, we choose the common Euclidean distance as basic density loss to measure the distance between predicted density map and ground truth. The original density loss is described as below:
\begin{equation}
\begin{aligned}
% L_{den} &= \frac{1}{N_b}\sum_{i=1}^{N_b}||G(x_i^s)-y_i^s||_2^2 \\
L_{den} = \frac{1}{N_b}\sum_{i=1}^{N_b}\sum_{k=1}^{N_\text{I}}((G(x_i^s))_k-y_{ik}^s)^2
\end{aligned}
\end{equation}
where $N_\text{I}$ is the total pixel number of the density map. With fine-grained scores, our final weighted density loss is:
\begin{equation}
\label{equ8}
\begin{aligned}
L_{dens}\! =\!\frac{1}{N_b}\sum_{i=1}^{N_b}(1\!+\!W_{1}^{i})\sum_{j=1}^{S^2}(1\!+\!W_{2}^{ij})\sum_{k=1}^{N_p^j}(1\!+\!W_{3}^{ijk})(1\!+\!W_{4}^{ijk})L_s
\end{aligned}
\end{equation}
where $N_p^j={N_\text{I}}/{S^2}$ is the pixel number of the $j$-th patch and $L_s = ((G(x_i^s))_{jk}\!-\!y_{ijk}^s)^2$. Here a residual mechanism is adopted during the weighting process for each score, which possesses more robustness to wrong discriminator output at the initial stage of network training. Finally, our proposed ASNet is trained end to end with the following loss:
\begin{equation}
{L_{All} = L_{dens}+\lambda_1 L_{adv1} + \lambda_2 L_{adv2}}
\end{equation}
The details of the overall training procedure can be seen in the supplementary.

% \subsection{Network Optimization}
% To train the full network parameters, including one generator $G$ and two discriminators $D_1$ and $D_2$, an alternative update is applied during the network optimization by iterative fixing the generator and two discriminators. Algorithm \ref{alg:A} describes the details of the overall training procedure. 

% \begin{algorithm}[t]
% \caption{Training procedure of the proposed adversarial scoring network (ASNet).}
% \label{alg:A}
% \begin{algorithmic}
% \REQUIRE source $X_s$, target $X_t$, generator $G(\cdot)$, global discriminator $D_1(\cdot)$ and local discriminator $D_2(\cdot)$
% \REQUIRE $\lambda_1$, $\lambda_2$, $\lambda_3$, training epochs $N$
% \FOR{$i\in [1,N]$}
% \FOR{$\text{minibatch $B^{(s)}$, $B^{(t)}$ $\in$ $X^{(s)}$, $X^{(t)}$}$ }
% \STATE \text{generate predicted density maps for both $B^{(s)}$ and $B^{(t)}$}\\
% \STATE \text{generate global discriminator map $O_1$ and $L_{d1}$ by $D_1$}\\
% \STATE \text{generate $W_1$, $W_3$} by $O_1$\\
% \STATE fix $G$, update $D_1$ by minimizing $L_{d1}$ \\
% \text{generate local discriminator map $O_2$ and $L_{d2}$ by $D_2$}\\
% \text{generate $W_2$, $W_4$} by $O_2$\\
% fix $G$, update $D_2$ by minimizing $\lambda_3L_{d2}$\\
% % \STATE \text{compute density loss $L_\text{den}$ for source images}
% \STATE \text{compute $L_{adv1}$ and $L_{adv2}$}\\
% \text{compute $L_{dens}$ by $W_1$, $W_3$, $W_2$, and $W_4$}\\
% {$L_{All} = L_{dens}+\lambda_1 L_{adv1} + \lambda_2 L_{adv2}$}\\
% fix $D_1,D_2$, update $G$ by minimizing $L_{All}$.
% \ENDFOR
% \ENDFOR
% \end{algorithmic}
% \end{algorithm}

\section{Experiments}
In this section, we provide a comprehensive evaluation of the proposed model on three adaptation experiments and a thorough ablation study to validate the key components of our algorithm. Experimental results demonstrate the effectiveness of our approach in domain adaptation for crowd counting. 

\subsection{Implement Details}
For a fair comparison, we use VGG-16 \cite{simonyan2014very} structure as the generator $G$. The final pooling layer and two fully connected layers are replaced by two dilated convolutional layers and a convolutional layer. The discriminator contains five convolutional layers with stride of 2 and kernel size 4 $\times$ 4, the channels of each layer are 64, 128, 256, 512, 1 respectively. Detailed configurations of the networks are shown in supplementary materials. The $G$ is trained using the Stochastic Gradient Descent (SGD) optimizer with a learning rate as $10^{-6}$. We use Adam optimizer \cite{kingma2014adam} with learning rate of $10^{-4}$ for the discriminators. During training, the $\lambda_1$, $\lambda_2$ and $\lambda_3$ are set to $10^{-3}$, $10^{-4}$, and $10^{-1}$ respectively. For data generation and augmentation, we follow the commonly used methods introduced in MCNN \cite{zhang2016single}. All input patches are resized to $512 \times 512$ with 3 channels.

\subsection{Datasets and Metric}
For the domain adaptation problem, we evaluate the proposed method on four publicly large-scale datasets, namely ShanghaiTech \cite{zhang2016single}, UCSD \cite{chan2008privacy}, Mall \cite{chen2012feature} and Trancos \cite{guerrero2015extremely} respectively.

\noindent\textbf{ShanghaiTech} consists of 1,198 annotated images with a total of 330,165 people with head center annotations. This dataset is divided into two parts: Part A with 482 images and Part B with 716 images. In Part A, all images are crawled from the Internet, of which 300 images are used for training and the rest are used for testing. Similarly, Part B is partitioned into 400 training images and 316 testing images captured from busy streets in Shanghai. Notably, Part A has density variations ranging from 33 to 3139 people per image with average count being 501.4 while Part B are relatively less diverse and sparser with an average density of 123.6.

\noindent\textbf{UCSD} is collected from a video camera at a pedestrian walkway, which contains a total of 49,885 pedestrian instances. This dataset is recorded by a video camera placed at a pedestrian walkway. It consists of 2000 frames, each of which has a resolution of 158 $\times$ 238. The region-of-interest (ROI) and
the perspective map are provided in the dataset. This dataset has a relatively low-density crowd since there are only 25 persons on average in each frame. Following \cite{chan2008privacy}, we use frames from 601 to 1,400 as the training set and the remaining frames for testing.

\noindent\textbf{Mall} is collected from a shopping mall by a public surveillance camera. This dataset contains 2,000 frames with diverse illumination conditions and crowd densities. Each frame has a fixed resolution of 320 $\times$ 240. In comparison to the UCSD dataset \cite{chan2008privacy}, this dataset has relatively higher crowd density images with an average count of 31 per image. We follow the pre-defined settings to use the
first 800 frames as the training set and the rest as the test set.

\noindent\textbf{Trancos} is a public vehicle dataset, which consists of 1244 images taken from traffic surveillance cameras located along different roads. The region-of-interest (ROI) is also provided for training and evaluation. Each vehicle is labeled with a single point annotation of its location. In total, there are 46,796 vehicle point annotations. There is a large discrepancy between Trancos and counting datasets. Different from counting dataset, Trancos is composed of multiple scenes but the same scenes appear in the training and test sets.

\noindent\textbf{Evaluation Metric} As commonly used in previous works, we adopt Mean Absolute Error (MAE) and Mean Squared Error (MSE) to evaluate the estimation performance of counting datasets. They are formulated 
as:
\begin{small}
\begin{equation}
    MAE =\frac{1}{N}\sum_{i=1}^{N}|C_{i}-\hat{C_{i}}|,
    MSE =\sqrt{\frac{1}{N}\sum_{i=1}^{N}(C_{i}-\hat{C_{i}})^{2}}
\end{equation}
\end{small}

\noindent where $N$ means numbers of image, $C_{i}$ means the total count of the estimated density map, and $\hat{C_{i}}$ refers to the total count of corresponding ground truth. Different for the vehicle dataset, we use the
Grid Average Mean absolute Error (GAME) metric, which is defined as:
\begin{equation}
    GAME(L)=\frac{1}{N}\sum_{i=1}^{N}(\sum_{l=1}^{4^{L}}|C_{i}^{l}-\hat{C_{i}^{l}}|)
\end{equation}
Given a specific number $L$, the $GAME(L)$ divides each image into $4^{L}$ non-overlapping regions of equal area, $C_{i}^{l}$ is the estimated count for image $i$ within region $l$, and $\hat{C_{i}^{l}}$ is the corresponding ground truth count. Note that $GMAE(0)$ is equivalent to $MAE$ metric. 

\subsection{Adaptation Results}
We provide a quantitative evaluation by performing
three sets of adaptation experiments: ShanghaiTech Part A $\rightarrow$ Part B, UCSD $\rightarrow$ Mall, and ShanghaiTech Part A $\rightarrow$ Trancos. For each pair of datasets, we report the errors between the generated density maps and the ground truth maps on the target set. We define several variants of the proposed model in the following: 1) \textbf{NoAdapt:} the model is only trained on the source samples. 2) \textbf{CoarseAdapt:} we perform the distribution alignment via a global discriminator and a local discriminator in an adversarial training scheme. 3) \textbf{FineAdapt:} the full model of our ASNet, which adds all the significance-aware scores to the CoarseAdapt. The list of methods to compare can be classified into four categories: 1) directly trained on the target data; 2) merely trained on the synthetic data (syn); 3) semi-supervised methods (semi); 4) merely trained on the real source data.

\begin{table}
    \centering
    \caption{The comparison results with previous methods for ShanghaiTech Part A $\rightarrow$ Part B. (TS: Target Supervision)}
    \label{shanghai}
    \begin{tabular}{p{3.0cm}<\centering|p{1.0cm}<\centering|p{1.0cm}<\centering p{1.0cm}<\centering}
    \toprule
    Method & TS & MAE & MSE \\ \midrule
    MCNN \cite{zhang2016single} & yes & 26.4 & 41.3  \\ 
    CP-CNN \cite{sindagi2017generating} & yes & 20.1 & 30.1\\ 
    IG-CNN \cite{sam2018divide} & yes & 13.6	& 21.1 \\ \midrule
    Cycle GAN \cite{zhu2017unpaired} & syn & 25.4  & 39.7 \\
    SE Cycle GAN \cite{wang2019learning} & syn & 19.9 & 28.3 \\
    SE+FD \cite{han2020focus} & syn & 16.9 & 24.7 \\ \midrule
    D-ConvNet-v1 \cite{shi2018crowd} & no & 49.1 & 99.2\\
    RegNet \cite{liu2019crowd} & no & 21.65 & 37.56 \\
    CODA \cite{li2019coda} & no &  15.9 & 26.9 \\\midrule
    NoAdapt (ours) & no & 27.28 & 35.14 \\
    CoarseAdapt (ours) & no & 15.77 & 24.92 \\
    FineAdapt (ours) & no & \textbf{13.59} & \textbf{23.15} \\
    \bottomrule
    \end{tabular}
\end{table}

First, we conduct the experiments about adapting ShanghaiTech Part A to Part B. As is shown in Table \ref{shanghai}, it is obvious that the proposed model outperforms existing domain adaptation methods by a large margin. In specific, our method improves MAE performance from 15.9 to 13.59. When comparing our method with \cite{zhu2017unpaired,wang2019learning,han2020focus}, which is merely trained on the much larger and more diverse synthetic dataset, we can achieve more superior results. Even compared with the mainstream supervised methods that are directly trained on the target domain, such as IG-CNN \cite{sam2018divide}, our model can still deliver competitive performance (MAE 13.59 vs 13.6). By observing the results gap between the NoAdapt and FineAdapt, we can find that ASNet yields a huge improvement after fine adaptation from 27.28/35.14 to 13.59/23.15.

\begin{table}
    \centering
    \caption{The comparison results with previous methods for UCSD $\rightarrow$ Mall. (TS: Target Supervision)}
    \label{mall}
    \begin{tabular}{p{3.0cm}<\centering|p{1.0cm}<\centering|p{1.0cm}<\centering p{1.0cm}<\centering}
    \toprule
    Method & TS & MAE & MSE \\ \midrule
    MORR \cite{chen2012feature} & yes & 3.15 & 15.7 \\
    ConvLSTM-nt \cite{xiong2017spatiotemporal} & yes & 2.53 & 11.2 \\ 
    MCNN \cite{zhang2016single} & yes & 2.24 & 8.5 \\ \midrule
    FA \cite{change2013semi} & semi & 7.47 & - \\
    HGP \cite{yu2005learning} & semi & 4.36 & -\\
    GPTL \cite{liu2015bayesian} & semi & 3.55 & - \\ \midrule
    CSRNet \cite{li2018csrnet} & no & 4.00 & 5.01 \\
    CODA \cite{li2019coda} & no & 3.38 & 4.15 \\
    \midrule
    NoAdapt (ours) & no & 4.19 &  5.03 \\
    CoarseAdapt (ours) & no & 3.47 &  4.12 \\
    FineAdapt (ours) & no & \textbf{2.76} & \textbf{3.55} \\
    \bottomrule
    \end{tabular}
\end{table}

Second, we use the UCSD as our source dataset and Mall as the target dataset. The results are shown in Table \ref{mall}. Obviously, our proposed method in an unsupervised setting outperforms all semi-supervised methods, which reduces the estimation errors by 22.2\% compared to the best semi-supervised model GPTL \cite{liu2015bayesian}. Our model can be gradually improved by incorporating different mechanisms. In specific, CoarseAdapt improves the MAE performance from 4.19 to 3.47 compared with NoAdapt, and FineAdapt further decreases the error to 2.76. Besides, our ASNet achieves the lowest MAE (the highest accuracy) compared to other domain adaptive methods \cite{li2019coda}. The above results demonstrate the effectiveness of our proposed adaptation pattern. 

\begin{table}
    \centering
    \caption{The comparison results with previous methods for ShanghaiTech Part A $\rightarrow$ Trancos. (TS: Target Supervision)}
    \label{trancos}
    \begin{tabular}{p{2.5cm}<\centering|c|p{0.85cm}<\centering p{0.85cm}<\centering p{0.85cm}<\centering p{0.85cm}<\centering}
    \toprule
    Method & TS & GAME0 & GAME1 & GAME2 & GAME3 \\ \midrule
    Lempitsky \cite{lempitsky2010learning} & yes & 13.76 & 16.72 & 20.72 & 24.36  \\
    Hydra 3s \cite{onoro2016towards} & yes & 10.99 & 13.75 & 16.69 & 19.32 \\ 
    AMDCN \cite{deb2018aggregated}   & yes &  9.77  & 13.16 & 15.00 & 15.87 \\
    FCNN-skip \cite{kang2018beyond}  & yes &  4.61 & 8.39 & 11.08 &16.10  \\ \midrule
    CSRNet \cite{li2018csrnet} & no & 13.71 & 13.81 & 14.52 & 15.75 \\
    CODA \cite{li2019coda} & no & 4.91 & 9.89 & 14.88 & 17.55 \\ \midrule
    NoAdapt (ours) & no & 13.78 & 13.83 & 15.32 &  15.88\\
    CoarseAdapt (ours) & no & 7.21 & 11.32 & 14.42 &  17.27 \\
    FineAdapt (ours) & no & \textbf{4.77} & \textbf{8.39} & \textbf{13.37} & \textbf{15.12}\\
    \bottomrule
    \end{tabular}
\end{table}

Third, we consider the experiments from ShanghaiTech Part A to Trancos, shown in Table \ref{trancos}. Distinctly, the proposed method yields an improvement of 2.9\% over the existing adaptation methods \cite{li2019coda}. Due to the large domain shift between the counting dataset and the vehicle dataset, we can see that the baseline (NoAdapt) fails to predict the density values for Trancos since there is little difference between GAME metrics. However, our model can reduce the estimation error from 13.78 to 4.77 close to the SOTA results, which proves the versatility of the proposed method.

\begin{figure*}[t]
\centering
\includegraphics[width=0.96\linewidth]{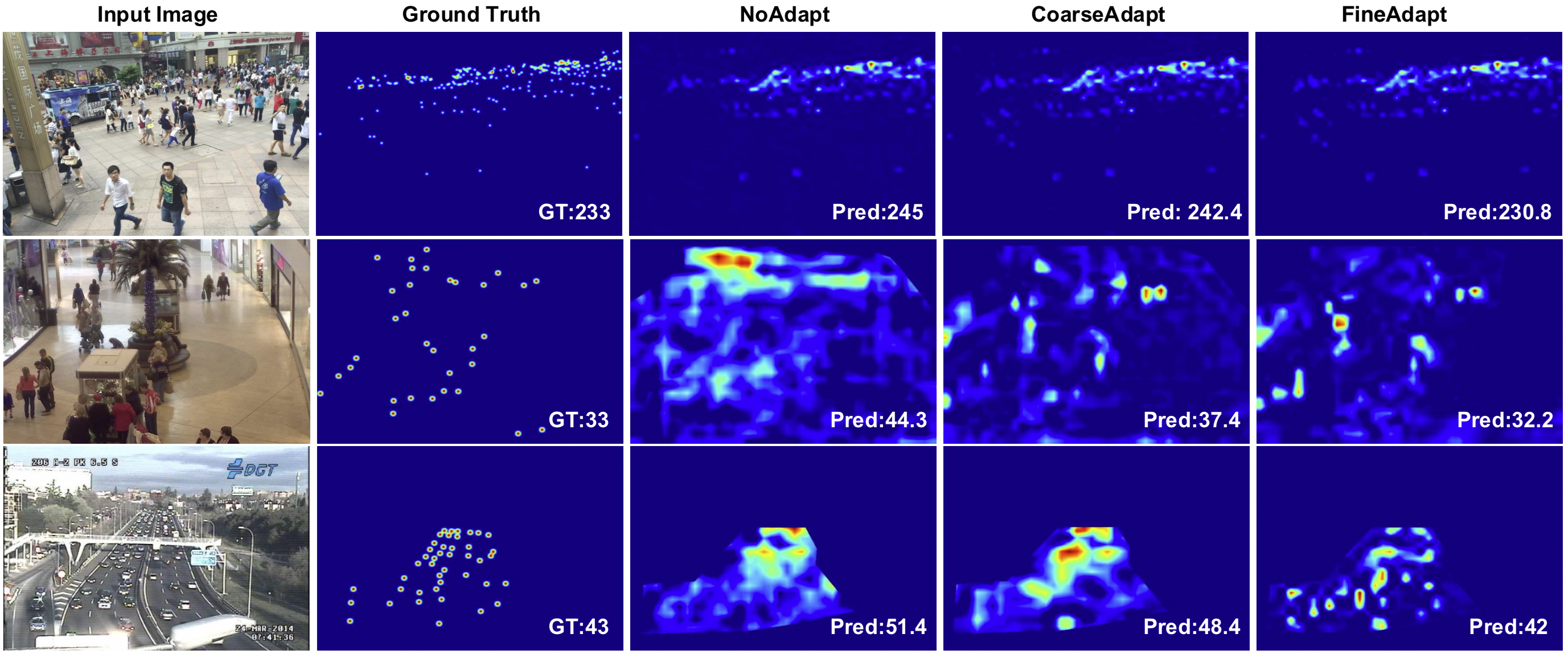}
\caption{Qualitative adaptation results. From top to down: ShanghaiTech Part B, Mall and Trancos, respectively.}
\label{visualization}
\end{figure*}

To better understand the superiority of the ASNet, we visualize the generated results of the step-wise variants in Figure \ref{visualization}. 
% From left to right, we show the input image, ground truth, NoAdapt, CoarseAdapt, and FineAdapt respectively. The first row presents the results of ShanghaiTech Part A $\rightarrow$ Part B, which demonstrates that FineAdapt predicts the crowd number more accurately than NoAdapt and CoarseAdapt. For the experiments from UCSD to Mall, NoAdapt can only reflect the crowd distribution trend while failing to locate each pedestrian. The reason for this is that the images of the UCSD dataset are gray-scale while the Mall dataset is composed of RGB images. After adaptation, the model has the ability to figure out the relatively accurate distribution of crowds. Since there is a large domain gap between the counting dataset and the vehicle dataset, models trained merely on ShanghaiTech Part A (NoAdapt) struggle to generate well on Trancos and produce meaningless density maps. It is worth mentioning that models with fine adaptation vastly promote the quality of the predicted density maps. In conclusion, the proposed methods can generate better results than other methods across domains. 
NoAdapt can only reflect the crowd distribution trend while failing to locate each pedestrian. After coarse adaptation, CoarseAdapt has the ability to figure out the relatively accurate distribution of crowds. It is obvious that FineAdapt vastly promote the quality of the predicted density maps. In conclusion, the proposed methods can generate better results than other methods across domains. More visualization results are shown in supplementary materials.

\subsection{Ablation Study}
In this section, we conduct abundant ablation experiments to analyze the components of the proposed ASNet. All ablations are conducted in the ShanghaiTech Part A $\rightarrow$ Part B setting for its large variations in crowd density.

\noindent\textbf{Effect of different components.} In this part, we analyze the effect of each component in the proposed method. From the results shown in Table \ref{components}, we can find the final performance has been gradually improved with the addition of each component, which illustrates the effectiveness of the proposed modules. To be specific, the errors are significantly reduced (MAE from 27.28 to 16.84) by only joining the global discriminator. When merely adding the local discriminator, the MAE errors are reduced to 19.12. Combining the two discriminators further optimizes the results to 15.77 MAE. These results indicate that image-level and patch-level alignment both play an important role in closing the data distribution across domains. Different levels of scores (image-level, patch-level, pixel-level, patch-pixel level) all contribute to the transferability of the model at different degrees, yielding MAE performance gains of 6.4\%, 7.5\%, 11.2\%, 13.8\% with the step-wise overlay of each score. All the above experimental results prove that our modules have a positive effect on each other, which is conductive to the accuracy of the adaptation.

\begin{table}[t]\small
\newcommand{\tabincell}[2]{\begin{tabular}{@{}#1@{}}#2\end{tabular}}
% \footnotesize
% \renewcommand{\arraystretch}{1.3}
\centering
\caption{Effects of different model components in ShanghaiTech Part A $\rightarrow$ Part B setting. \textit{G-D} and \textit{L-D} mean the global and local discriminator, \{\textit{I}, \textit{P}, \textit{PI}, \textit{P-PI}\} correspond to \{image, patch, pixel, patch-pixel\} level scores respectively.}
\label{components}
\begin{tabular}{p{1.0cm}<\centering|cc|p{0.2cm}<\centering p{0.2cm}<\centering p{0.2cm}<\centering p{0.7cm}<\centering|p{0.5cm}<\centering p{0.5cm}<\centering}
\toprule
\multirow{2}{*}{NoAdapt} & \multicolumn{2}{c|}{CoarseAdapt} & \multicolumn{4}{c|}{FineAdapt} & \multirow{2}{*}{MAE} & \multirow{2}{*}{MSE} \\
\cline{2-7}
& \textit{G-D} & \textit{L-D} & \textit{I} & \textit{P} & \textit{PI} & \textit{P-PI} & & \\
\toprule
\checkmark & & & & & & & 27.28 & 35.14\\
& \checkmark & & & & & & 16.84 & 27.37\\ 
& & \checkmark & & & & & 19.12 & 30.02\\
& \checkmark & \checkmark & & & & & 15.77 & 24.92 \\ 
& \checkmark & \checkmark & \checkmark & & & & 14.76 & 24.18\\ 
& \checkmark & \checkmark & \checkmark & \checkmark& & & 14.58& 24.15\\ 
& \checkmark & \checkmark & \checkmark & \checkmark& \checkmark & & 14.01 & 24.02\\ 
& \checkmark & \checkmark & \checkmark & \checkmark& \checkmark& \checkmark& \textbf{13.59} & \textbf{23.15}\\ 
\bottomrule
\end{tabular}
\end{table}

% \begin{table}[h]
%     \centering
%     \caption{Ablation study on the residual mechanism.}
%     \label{residual}
%     \begin{tabular}{c|cc}
%     \toprule
%     Method & MAE & MSE \\ \midrule
% ASNet w/o residual mechanism & 14.67 & 24.22 \\
% ASNet w/ residual mechanism & \textbf{13.59} & \textbf{23.15} \\
%     \bottomrule
%     \end{tabular}
% \end{table}

% \noindent\textbf{Effect of residual mechanism.} In this section, we evaluate the effectiveness of the residual mechanism used in the equation \ref{equ8}. This strategy is designed to avoid the inaccurate discriminator output at the initial state of the network optimization, thus improving the robustness of the scores guidance. The experimental results are summarized in Table \ref{residual}. We can observe that the performance of the proposed ASNet improves from MAE 14.67 to 13.59 as the residual mechanism is adopted. These experiments well demonstrate the effectiveness of our designed residual mechanism.

\noindent\textbf{Effect of global and local modules.} In this section, we separately study the impact of global and local modules on the final model performance. We split the model components into two categories: global-related modules (global discriminator, image and pixel level scores) and local-related modules (local discriminator, patch and patch-pixel level scores). As is illustrated in Table \ref{global_local}, the global-related modules significantly boost the performance from MAE 27.28 to 14.89 since they fully reduce the domain shift among the source and the target domains in a global view. Also, the local-related modules degrade the estimations errors to MAE 15.13. This proves that utilizing the patches to close the domain gap is still effective. The full model achieves the best performance with respect to MAE and MSE, which demonstrates that global-related and local-related modules mutually refine each other and thus contribute together to the final performance of the proposed ASNet.

\begin{table}[t]
    \centering
    \caption{Ablation study on the global and local modules.}
    \label{global_local}
    \begin{tabular}{p{3.5cm}<\centering|p{0.85cm}<\centering p{0.85cm}<\centering}
    \toprule
    Method & MAE & MSE \\ \midrule
NoAdapt & 27.28 & 35.14 \\
Global-related modules & 14.89 & 24.36 \\
Local-related modules & 15.13 & 24.48 \\
FineAdapt (full model) & \textbf{13.59} & \textbf{23.15} \\
    \bottomrule
    \end{tabular}
\end{table}

\begin{figure*}[t]
\centering
\includegraphics[width=\linewidth]{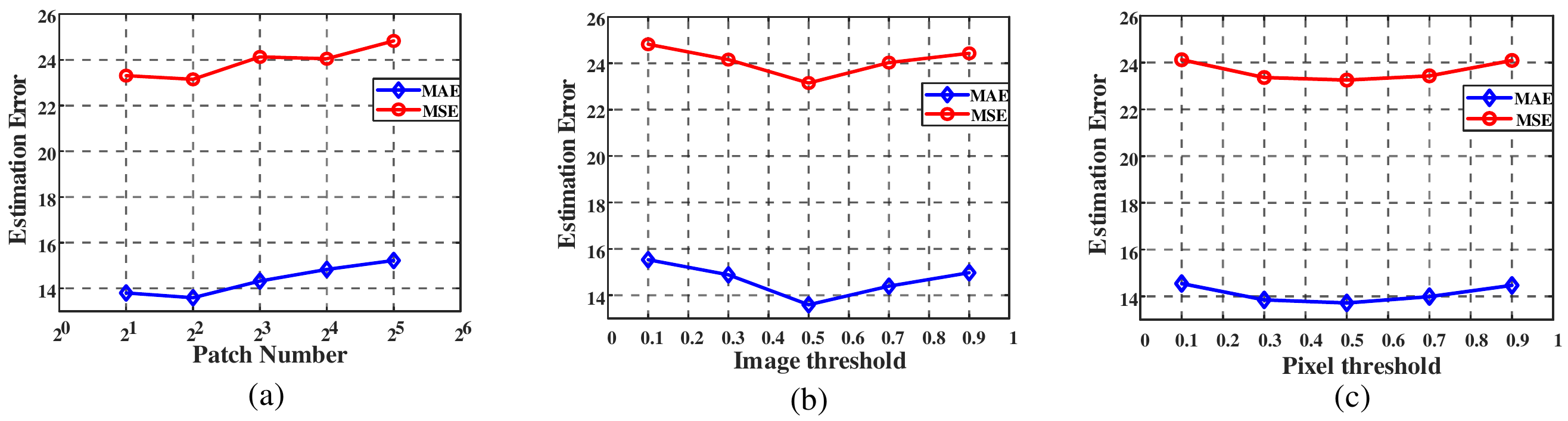}
\caption{Effect of using different patch number $S$ (total patches number is ${S}^2$), image-level threshold, and pixel-level threshold for the whole training. Here we do not show the soft threshold value in the image (C).}
\label{method}
\end{figure*}

\begin{figure*}[t]
\centering
\includegraphics[width=\linewidth]{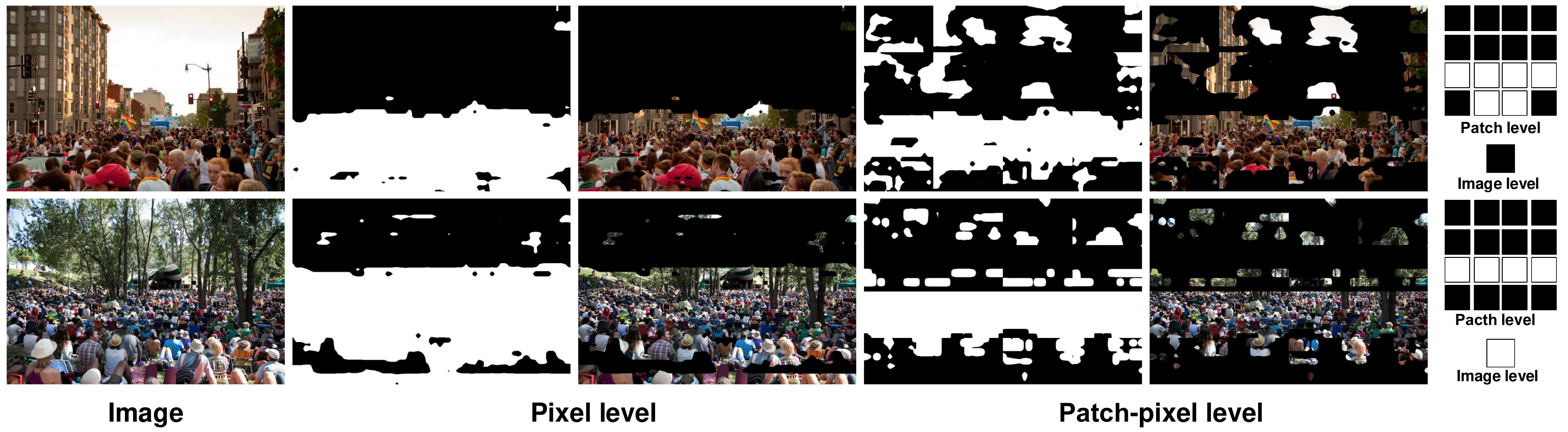}
\caption{Visualization of different levels (pixel, patch-pixel, patch, image respectively) level scores generated by the dual-discriminator. A square in the figure represents a scalar. Note white square refers to 1 while the black square refers to 0.}
\label{weight}
\end{figure*}

\noindent\textbf{Effect of patch number.} In our proposed method, we divide the density map into $S \times S$ patches and send them to the local discriminator for the subsequent steps. We evaluate how the patch number $S$ affects the final performance in this part. As shown in Figure \ref{method} (a), the results of the model are robust to the patch numbers. However, when the patch number is too large, the complexity of calculation increases with a slight decrease in the estimation performance. To achieve the best accuracy, we set the patch number $S$ to 4 throughout the experiments.

\noindent\textbf{Effect of image-level threshold.} Figure \ref{method} (b) illustrates the impact of the image-level threshold on the performance of the proposed model. Obviously, when the threshold is too small or too large, the result turns out a cliff-like decline. The main reason is that too small threshold filters out mistakenly a mass of similar images in the source domain while too large threshold may introduce some dissimilar samples. It can be observed that threshold = 0.5 achieves the best result, so we use this value throughout the experiments.

\noindent\textbf{Effect of pixel-level threshold.} Figure \ref{method} (c) illustrates the impact of changes in pixel-level threshold on the final performance of the proposed model. We change the soft threshold into different hard thresholds. Obviously, the estimation error reaches the minimum when threshold = 0.5. Note this result is still inferior to adopting a soft threshold, which demonstrates the rationality of our choice.

\noindent\textbf{Visualization of weight maps.} We show the scores of different levels generated by the dual-discriminator in Figure \ref{weight}. For the pixel level, we can see that the probability maps focus on where the crowds distribute. This indicates that the regions containing crowds are transferable while unique scene attributes such as background objects are disturbing noise. Besides, since pixel level scores provide a global view while patch level scores pay attention to local information of patches, these two maps could be complementary to each other. The image-level scores determine the transferability of the entire image and the patch level scores illustrate the trade-off of corresponding patches. These results intuitively show that our model can generate reasonable scores for the fine-grained knowledge transfer across domains.  

\section{Conclusion and Future work}
In this paper, we propose a novel adversarial scoring network (ASNet) for domain adaptive crowd counting. Unlike previous methods, our model can adaptively select the transferable regions or images to achieve the fine-grained knowledge transfer across domains. To implement this goal, we design a dual-discriminator strategy to conduct a coarse distribution alignment and generate significance-aware scores of different levels based on the transferability of source samples. With these scores as a signal to guide the density optimization, our model can better mitigate the domain gap at multiple perspectives, thus significantly boosting the adaptation accuracy. Three sets of adaptation experiments and thorough ablation studies demonstrate the effectiveness of our proposed method. To further verify the effectiveness of our method, we evaluate the situation from synthetic datasets to the real-world Shanghaitech datasets and compare our model with the latest unsupervised methods. The detailed results can be seen in the supplementary material due to the limited space.

In the future, we are interested to evaluate more situations in counting area and extend our method to other tasks such as object detection and depth estimation.

\begin{acks}
This work is supported by National Natural Science Foundation of China (NSFC) under grant no. 61972448. (Corresponding author: Pan Zhou.
\end{acks}

\appendix
\section{Synthetic dataset}
In this section, we conduct the experiments about adapting GCC dataset to ShanghaiTech Part B. As is shown in Table \ref{gcc}, it is obvious that the proposed model outperforms existing domain adaptation methods by a large margin.

\begin{table}[h]
    \centering
    \caption{The comparison results with previous methods for GCC $\rightarrow$ ShanghaiTech Part B.}
    \label{gcc}
    \begin{tabular}{p{3.0cm}<\centering|p{1.0cm}<\centering|p{1.0cm}<\centering p{1.0cm}<\centering}
    \toprule
    Method & MAE & MSE \\
    \midrule
    CycleGAN \cite{zhu2017unpaired} & 25.4 & 39.7 \\
    SE CycleGAN \cite{wang2019learning} & 19.9 & 28.3 \\
    SFCN+MFA+SDA \cite{gao2019feature} & 16.0 & 24.7 \\
    SE+FD \cite{han2020focus} & 16.9 & 24.7 \\
    SE CycleGAN (JT) \cite{wang2021pixel} & 16.4 & 25.8 \\
    ASNet (ours) & \textbf{14.6} & \textbf{22.6} \\
    \bottomrule
    \end{tabular}
\end{table}

\section{Residual Mechanism}
In this section, we evaluate the effectiveness of the residual mechanism used in the Equation 8. This strategy is designed to avoid the inaccurate discriminator output at the initial state of the network optimization, thus improving the robustness of the scores guidance. The experimental results are summarized in Table \ref{residual}. We can observe that the performance of the proposed ASNet improves from MAE 14.67 to 13.59 as the residual mechanism is adopted. These experiments well demonstrate the effectiveness of our designed residual mechanism.
\begin{table}[h]
    \centering
    \caption{Ablation study on the residual mechanism.}
    \label{residual}
    \begin{tabular}{c|cc}
    \toprule
    Method & MAE & MSE \\ \midrule
ASNet w/o residual mechanism & 14.67 & 24.22 \\
ASNet w/ residual mechanism & \textbf{13.59} & \textbf{23.15} \\
    \bottomrule
    \end{tabular}
\end{table}

\begin{table}[h]
    \large
    \centering
    \captionsetup{font={large}}
    \setlength{\abovecaptionskip}{5pt}  
    \caption{\large{The architecture of the generator.}}
    \setlength{\tabcolsep}{7mm}
    \begin{tabular}{c}
    \toprule[1.5pt]
    \textbf{\textit{G}} \\
    \midrule[1.5pt]
    \textbf{Convolution Layers} \\
    $\left[ \textnormal{K(3,3)-c64-s1-p1-R} \right]$ $\times$ 2 \\
    Max pooling \\
    $\left[ \textnormal{K(3,3)-c128-s1-p1-R} \right]$ $\times$ 2 \\
    Max pooling \\
    $\left[ \textnormal{K(3,3)-c256-s1-p1-R} \right]$ $\times$ 3 \\
    Max pooling \\
    $\left[ \textnormal{K(3,3)-c512-s1-p1-R} \right]$ $\times$ 3 \\
    Max pooling \\
    $\left[ \textnormal{K(3,3)-c512-s1-p1-R} \right]$ $\times$ 3 \\
    \midrule[1pt]
    \textbf{Dilation Layers} \\
    K(3,3)-c256-s1-p4-d4-R \\
    K(3,3)-c64-s1-p4-d4-R \\
    \midrule[1pt]
    \textbf{Output Layer} \\
    K(3,3)-c1-s1-p1-d1 \\
    \bottomrule[1.5pt]
    \end{tabular}
    \label{generator}
\end{table}

\section{Detailed Architecture}
\label{sec:architecture}
In this section, we introduce the detailed structure of each component in our adversarial scoring network. Table \ref{generator} illustrates the configuration of the generator. For fair comparisons with previous methods, we use VGG-16 structure as the generator $G$.  In the table, “k(3,3)-c256-s1-p4-d4-R” denotes the convolutional operation with kernel size of 3 $\times$ 3, 256 output channels, stride size of 1, padding size of 4, dilation size of 4, and 'R' means the ReLU layer. Table \ref{discriminator} explains the architecture of the dual-discriminator, where 'LR' indicates the leaky ReLU layer. 

% Each discriminator contains five convolutional layers with stride of 2 and kernel size 4 $\times$ 4, the channels of each layer are 64, 128, 256, 512, 1 respectively. Note that the global discriminator $D_1$ and local discriminator $D_2$ have the same network structure but accept different inputs.

\begin{table}[h]
    \large
    \centering
    \captionsetup{font={large}}
    \setlength{\abovecaptionskip}{5pt}
    \caption{\large{The architecture of the dual-discriminator.}}
    \setlength{\tabcolsep}{8mm}
    \begin{tabular}{c}
    \toprule[1.5pt]
    \textbf{\textit{${D_i (i=1,2)}$}}\\
    \midrule[1.5pt]
    \textbf{Convolution Layers} \\
    K(4,4)-c64-s2-p1-LR \\[2pt]
    K(4,4)-c128-s2-p1-LR \\[2pt]
    K(4,4)-c256-s2-p1-LR \\[2pt]
    K(4,4)-c512-s2-p1-LR \\[2pt]
    K(4,4)-c1-s1-p2 \\[2pt]
    \midrule[1pt]
    \textbf{Activation Layer} \\
    Sigmoid \\
    \bottomrule[1.5pt]
    \end{tabular}
    \label{discriminator}
\end{table}

\section{Algorithm}
To train the full network parameters, including one generator $G$ and two discriminators $D_1$ and $D_2$, an alternative update is applied during the network optimization by iterative fixing the generator and two discriminators. Algorithm \ref{alg:A} describes the details of the overall training procedure.

\begin{algorithm}[h]
\caption{Training procedure of the proposed ASNet.}
\label{alg:A}
\begin{algorithmic}
\REQUIRE source $X_s$, target $X_t$, generator $G(\cdot)$, global discriminator $D_1(\cdot)$ and local discriminator $D_2(\cdot)$
\REQUIRE $\lambda_1$, $\lambda_2$, $\lambda_3$, training epochs $N$
\FOR{$i\in [1,N]$}
\FOR{$\text{minibatch $B^{(s)}$, $B^{(t)}$ $\in$ $X^{(s)}$, $X^{(t)}$}$ }
\STATE \text{generate predicted density maps for both $B^{(s)}$ and $B^{(t)}$}\\
\STATE \text{generate global discriminator map $O_1$ and $L_{d1}$ by $D_1$}\\
\STATE \text{generate $W_1$, $W_3$} by $O_1$\\
\STATE fix $G$, update $D_1$ by minimizing $L_{d1}$ \\
\text{generate local discriminator map $O_2$ and $L_{d2}$ by $D_2$}\\
\text{generate $W_2$, $W_4$} by $O_2$\\
fix $G$, update $D_2$ by minimizing $\lambda_3L_{d2}$\\
% \STATE \text{compute density loss $L_\text{den}$ for source images}
\STATE \text{compute $L_{adv1}$ and $L_{adv2}$}\\
\text{compute $L_{dens}$ by $W_1$, $W_3$, $W_2$, and $W_4$}\\
{$L_{All} = L_{dens}+\lambda_1 L_{adv1} + \lambda_2 L_{adv2}$}\\
fix $D_1,D_2$, update $G$ by minimizing $L_{All}$.
\ENDFOR
\ENDFOR
\end{algorithmic}
\end{algorithm}
\clearpage
\vfill\eject
\bibliographystyle{ACM-Reference-Format}
\bibliography{sample-base}


\end{document}